\pdfoutput=1

\documentclass[11pt]{article}

\usepackage[preprint]{acl}

\usepackage{times}
\usepackage{latexsym}

\usepackage[T1]{fontenc}

\usepackage[utf8]{inputenc}

\usepackage{microtype}

\usepackage{inconsolata}

\usepackage{graphicx}
\usepackage{utfsym}
\usepackage{booktabs}  
\usepackage{xparse}
\usepackage{amsmath} 
\usepackage{amssymb}
\usepackage[utf8]{inputenc}
\usepackage{tcolorbox}
\tcbuselibrary{listingsutf8,skins} 
\usepackage{makecell}

\def \taken {$^\dagger$}
\def \loss {\mathcal{L}}
\def \h {\mathbf{h}}
\def \a {\mathbf{a}}
\def \q {\mathbf{q}}
\def \st {\texttt{<st>}}
\def \R {\mathbb{R}}

\newcommand\norm[1]{\left\| #1 \right\|}

\newcommand\sbr[1]{\left( #1 \right)}
\newcommand\mbr[1]{\left[ #1 \right]}
\newcommand\bbr[1]{\left\{ #1 \right\}}
\newcommand\logcond[2]{\log \sbr{#1 \mid #2}}

\newcommand\example[4]{
    \begin{tcolorbox}[
        colback=gray!20,       
        colframe=black!80,     
        sharp corners=all,     
        boxrule=0.8pt,         
        title=#1,           
        fonttitle=\bfseries,   
        coltitle=white,        
        colbacktitle=black!80, 
        enhanced,              
        left=4mm,              
        right=4mm,             
        top=2mm,               
        bottom=2mm             
    ]
    Question = #2

    \if\relax\detokenize{#3}\relax
    \else
        CoT = #3
    \fi

    Answer = #4
    \end{tcolorbox}
}

\NewDocumentCommand{\avg}{o m}{%
  \frac{1}{#2} \sum_{\IfValueTF{#1}{#1}{i}=1}^{#2}%
}

\title{DART: Distilling Autoregressive Reasoning to Silent Thought}

\author{
 \textbf{Nan~Jiang\textsuperscript{1,2}\thanks{These authors contributed equally to this work}\thanks{This work was done during the internship at Tencent Inc.}},
 \textbf{Ziming~Wu\textsuperscript{2}\footnotemark[1]\thanks{Corresponding author.}},
 \textbf{De-Chuan~Zhan\textsuperscript{1}},
 \textbf{Fuming Lai\textsuperscript{2}},
 \textbf{Shaobing Lian\textsuperscript{2}}
\\
 \textsuperscript{1}Nanjing University, 
 \textsuperscript{2}Tencent Inc.
\\
\small{
 \texttt{jiangn@lamda.nju.edu.cn, \{jimmyzmwu, fuminglai, lokilian\}@tencent.com, zhandc@nju.edu.cn} 
}
}

\begin{document}
\maketitle
\begin{abstract}

Chain-of-Thought (CoT) reasoning has significantly advanced Large Language Models (LLMs) in solving complex tasks. 
However, its autoregressive paradigm leads to significant computational overhead, hindering its deployment in latency-sensitive applications. 
To address this, we propose \textbf{DART} (\textbf{D}istilling \textbf{A}utoregressive \textbf{R}easoning to Silent \textbf{T}hought), a self-distillation framework that enables LLMs to replace autoregressive CoT with non-autoregressive Silent Thought (ST). Specifically, DART introduces two training pathways: the CoT pathway for traditional reasoning and the ST pathway for generating answers directly from a few ST tokens. The ST pathway utilizes a lightweight Reasoning Evolvement Module (REM) to align its hidden states with the CoT pathway, enabling the ST tokens to evolve into informative embeddings. During inference, only the ST pathway is activated, leveraging evolving ST tokens to deliver the answer directly. Extensive experimental results demonstrate that DART offers significant performance gains compared with existing non-autoregressive baselines without extra inference latency, serving as a feasible alternative for efficient reasoning.
\end{abstract}

\section{Introduction}

Large Language Models (LLMs) have demonstrated remarkable performance~\citep{deepseek2025r1, openai202501} across various reasoning tasks by leveraging Chain-of-Thought (CoT)~\citep{wei2022cot}, which decomposes complex problems into intermediate reasoning steps.
Despite these successes, the autoregressive nature of CoT introduces substantial computational cost, resulting in increased latency and limiting its effectiveness in real-time applications~\citep{sui2025stop}.

To alleviate this computational burden, implicit CoT reasoning~\citep{deng2023implicit,deng2024icot} performs implicit reasoning in the hidden state rather than the explicit CoT tokens to avoid extra computation. Continuous thought methods~\citep{hao2024coconut,cheng2024compressed} compress discrete textual tokens into compact, continuous representations, reducing the number of intermediate tokens without obvious degradation in reasoning capability. However, these existing approaches either suffer from unsatisfactory performance or remain haunted by the autoregressive generation paradigm, leading to suboptimal efficiency.

\begin{figure*}[t]
  \includegraphics[width=0.98\linewidth]{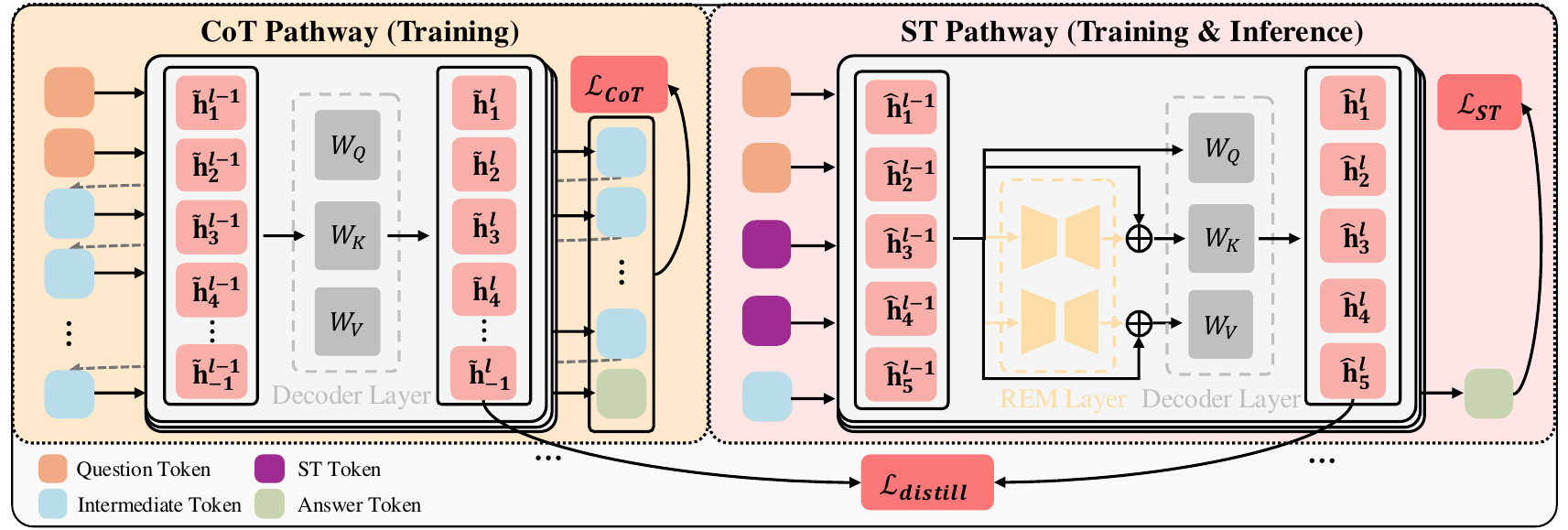}
  \caption {Overall Framework of DART. During inference, we employ the ST pathway to respond directly without step-by-step reasoning in prior work~\citep{wei2022cot, hao2024coconut}. The shared intermediate token represents the separator token. Feed-forward layer in the decoder layer is omitted for simplicity.
  }
  \label{fig:framework}
\end{figure*}

To this end, we propose DART (Distilling Autoregressive Reasoning to Silent Thought), a novel framework that enables the LLMs to internalize the autoregressive CoT into non-autoregressive Silent Thought (ST) with an excellent efficiency-efficacy trade-off. To be specific, DART employs two pathways in the training procedure as shown in Figure~\ref{fig:framework}, namely: the CoT pathway, which generates both the answer tokens and the explicit CoT tokens; and the ST pathway, which focuses solely on generating answers, conditioned on the ST tokens concatenated after the question. Additionally, the ST pathway introduces a lightweight Reasoning Evolvement Module (REM) to align the hidden state of the last word preceding the answer with that of the CoT pathway. During inference, initial ST tokens are appended to user input and processed through the REM-equipped ST pathway. Analogous to human cognition that progresses from vague conceptual abstraction to concrete resolution, these ST tokens evolve into increasingly informative embeddings as they propagate through the network, ultimately serving as a context-aware bridge between the instruction and its logically grounded response. Empirical results demonstrate that DART achieves significant efficiency gains while maintaining comparable performance. To summarize, our contributions are as follows:

\begin{itemize}
    \item We explore non-autoregressive ST as a promising alternative to the CoT paradigm, providing valuable insights for future work;
    \item We introduce DART, a simple but effective framework that employs REM to align autoregressive CoT with non-autoregressive ST in a dual-pathway architecture;
    \item We conduct extensive experiments to validate DART on multiple reasoning benchmarks, demonstrating its remarkable efficiency alongside satisfactory accuracy and interpretability.
\end{itemize}

\section{Related Work}

Empirical results and theoretical analysis~\citep{feng2023towards,liu2024chain} have demonstrated the effectiveness of CoT developed from supervised fine-tuning ~\citep{yue2024mammoth,yu2024metamath} and reinforcement learning~\cite{wang2024math,shao2024deepseekmath,deepseek2025r1}. However, intermediate steps in the CoT reasoning will cause extra computational cost, resulting in low throughput. To reduce this computation overhead, Coconut~\citep{hao2024coconut} employs curriculum learning to fine-tune an LLM capable of autoregressively generating final-layer hidden states to serve as the replacement of CoT tokens. These final-layer hidden states, dubbed continuous thought, are more information-dense, thus reducing the intermediate steps. One contemporaneous work, CODI~\cite{shen2025codi} also exploits the continuous thought but employs an end-to-end distillation framework rather than the curriculum learning. Despite the impressive performance of these methods, their efficiency is still limited by the autoregressive pattern. On the other hand, iCoT~\cite{deng2023implicit, deng2024icot} manages to embed the CoT reasoning within the model's hidden space. However, it lacks scalability for the larger models~\citep{shen2025codi}.
\section{Method}

\subsection{Dual-Pathway Architecture}

Given a question $Q$, our goal is to fine-tune a causal decoder-only LLM parameterized by $\theta$ to provide the proper answer $Y=\bbr{y_i}_{i=1}^{M}$. In DART, we introduce a dual-pathway architecture to allows two distinct answering ways during training.

\textbf{Chain-of-Thought Pathway.} This pathway adheres to the conventional CoT approach, where the model first produces a sequence of intermediate reasoning steps $Z=\bbr{z_i}_{i=1}^{N}$ before producing the final answer. During training, the cross-entropy loss for next token prediction is adopted for optimizing this pathway:
\begin{align*}
    \label{eq:loss:cot}
    \loss_{CoT} = 
    & -\avg{N}\logcond{z_i}{Q,z_{1:i-1};\theta} \\
    & -\avg{M}\logcond{y_i}{Q,Z,y_{1:i-1};\theta}.
\end{align*}
Notably, the first $t-1$ tokens of $Z$ are indeed CoT tokens, while the remaining are separator tokens shared with the ST pathway. In this paper, we fix $z_{t:N}$ as the answer prompt "\textit{Answer:}". 

\textbf{Silent Thought Pathway.} In contrast, the ST pathway directly generates the answer conditioned on the preceding ST sequence $S=\bbr{s_i}_{i=1}^{C}$ and separators $z_{t:N}$. Here, each $s_i$ is a special token \st~and $C$ is set as 20 in this paper. The objective function of this pathway can be formulated as
\begin{equation*}
    \loss_{ST}=-\avg{M}\logcond{y_i}{Q,X,y_{1:i-1};\theta,\phi},
    \label{eq:loss:st}
\end{equation*}
where $X=[S;z_{t:N}]$ and $\phi$ is the parameters associated with REM to be detailed in Section~\ref{sec:rem-distill}.

\subsection{REM-based Self-Distillation} 
\label{sec:rem-distill}

Our preliminary experiments show that enabling the evolution of the ST token requires more fine-grained supervision from CoT data to capture deeper intrinsic reasoning patterns. As revealed by the prior work~\citep{dai2023why}, the intermediate words essentially impose a shift to the hidden state of the last word before the answer. We can approximate this effect at the $l$-th decoder layer as
\begin{align*}
    \tilde{\a}^l & \approx \a^l + W_V^lH_{Z}^{l-1}(W_K^lH_{Z}^{l-1})^T\q^l, \\
    \tilde{\h}^l 
    & \approx {\h}^l + f\sbr{W_V^lH_{Z}^{l-1}(W_K^lH_{Z}^{l-1})^T\q^l},
\end{align*}
where $f(\cdot)$ denotes the feed-forward layer; $\q^l$ is the attention query vector of $z_{N}$ in $l$-th decoder layer; $\a^l$ and $\h^l$ indicate the output of the attention head and the output hidden state, given the question-only input; $H_{Z}^{{l-1}}$ represents the input hidden state of intermediate token sequence $Z$; and $W_K^l$, $W_V^l$ are the key and value projection matrices. A detailed derivation is provided in the Appendix~\ref{sec:appendix:derivation}.

Since the intermediate tokens are autoregressively generated conditioned on the question $Q$ and the model parameters $\theta$, the induced shift can be viewed as $g_{\theta^{1:l}}\sbr{h_{\theta}\sbr{Q}}$ where $Z=h_{\theta}(Q)$ and $\theta^{1:l}$ denotes the parameter of the first $l$ layers. Given that flattening the function $h_{\theta}(\cdot)$ is non-trivial, we propose to approximate the process by introducing a lightweight REM module at each decoder layer, which also leverages both the parameters $\theta$ and the in-context information from $Q$. Specifically, to induce such a shift, REM adapts the standard attention mechanism as:
\begin{equation*}
    \widehat{\a} \approx W_V \bar{W}_R^V[H_Q;H_X]\sbr{W_K\bar{W}_R^K[H_Q;H_X]}^T\q,
\end{equation*}
\begin{equation*}
    \widehat{\h} \approx {\h} + g_{\theta^{1:l},\phi^{1:l}}([Q;X]).
\end{equation*}
where $\bar{W}_R^J=\frac{\alpha}{d}W_{R_2}^J{W_{R_1}^J}^T + I$ for $J\in\bbr{K,V}$. Here, $W_{R_1}^J,W_{R_2}^J\in \R^{n \times d}$ are learnable matrices injected before the key and value projection matrices; $n,d$ are the hidden state dimension and the REM projection space dimension; and $\alpha$ is a scaling hyperparameter.
The superscript for the layer index is omitted for simplicity. 
REM offers two key advantages: (1) It introduces a few additional parameters while enabling rich interactions between $Q$ and $\theta$, effectively capturing contextual and domain-specific knowledge; (2) It is a simple plug-in module compatible with any decoder-only LLM, which can be seamlessly merged into the original architecture without increasing inference-time parameters.

Based on the analysis, we adopt the following distillation loss to guide the learning process: 
\begin{equation*}
    \loss_{distill}=\avg[l]{L}\frac{1}{\sigma(\tilde{\h}^l)}\norm{\tilde{\h}^l - \widehat{\h}^l}_1,
    \label{eq:loss:distill}
\end{equation*}
where $\sigma\sbr{\cdot}$ denotes the standard deviation within a batch. By aligning these hidden states, the function $g_{\theta^{1:l},\phi^{1:l}}([Q;X])$ is encouraged to approximate $g_{\theta^{1:l}}\sbr{h_{\theta}\sbr{Q}}$, thereby distilling the reasoning capability from the CoT pathway into the ST pathway. Furthermore, we empirically show in Section~\ref{sec:further_analysis} that the initial meaningless token \st~will evolve into an informative latent representation as it goes through the REM-equipped ST pathway, simulating a blur-to-concrete thinking process. To summarize, the overall objective function of DART is
\begin{equation*}
    \loss_{DART} = \loss_{CoT} + \loss_{ST} + \lambda \loss_{distll},
\end{equation*}
where $\lambda=20$ is a trade-off hyperparameter.

\begin{table*}[!t]
    \centering
    \resizebox{1.0\linewidth}{!}{
    \begin{tabular}{lc c lc c lc c lc c lc}
        \toprule
         &&& \multicolumn{2}{c}{In-Distribution} && \multicolumn{8}{c}{Out-of-Distribution} \\
        \cmidrule{4-5} \cmidrule{7-14}
        Methods & Is NAR? && \multicolumn{2}{c}{GSM8K} && \multicolumn{2}{c}{GSM-HARD} && \multicolumn{2}{c}{SVAMP} && \multicolumn{2}{c}{MultiArith}\\
        \midrule
        \noalign{\smallskip}
        CoT
        & \usym{2717}
        && 58.8 & 425
        && 13.4 & 477 
        && 59.9 & 227
        && 97.2 & 218 \\
        Coconut~\citep{hao2024coconut}
        & \usym{2717}
        && 50.6 & 390  
        && 11.2 & 483 
        && 53.1 & 181
        && 96.5 & 217 \\
        CODI~\citep{shen2025codi} 
        & \usym{2717}
        && 55.6\taken & 143   
        && 12.8\taken & 153 
        && 61.1\taken & 141
        && 96.1\taken & 132 \\
        \midrule
        No-CoT
        & \usym{2713}
        && 32.5 & 36
        && 7.1  & 57 
        && 40.6 & 34
        && 61.3 & 33 \\
        iCoT~\citep{deng2024icot} 
        & \usym{2713}
        && 19.0\taken & 36
        && 4.4 \taken & 57
        && 40.9\taken & 34
        && 39.0\taken & 33 \\
        PauseFT~\citep{Goyal2024pause}
        & \usym{2713}
        && 32.1 & 37 
        && 7.3  & 60
        && 40.4 & 35
        && 59.2 & 33 \\
        DART (Ours)
        & \usym{2713}
        && 42.6 & 37  
        && 10.9 & 60
        && 50.5 & 35
        && 84.8 & 33 \\
        \bottomrule
    \end{tabular}
    }
    \caption{Results on GSM8K, GSM-HARD, SVAMP, and MultiArith. Accuracy (\%) is on the left and inference time (ms) on the right for each benchmark.
    NAR stands for non-autoregressive.
    \taken The result is from \citep{shen2025codi}.
    }
    \label{tb:main_res}
\end{table*}
\begin{table}[!htbp]
    \centering
    \resizebox{1.0\linewidth}{!}{
    \begin{tabular}{l c c}
        \toprule
        Methods & FLOPs         & peak GPU memory \\
        \midrule
        CoT     & 1874774802    & 12.28 \\
        No-CoT  & 1398731734    & 11.70 \\
        DART    & 1881308345    & 18.07 \\

        \bottomrule
    \end{tabular}
    }
    \caption{Total FLOPs (GF) and peak GPU memory (GB) for CoT, No-CoT, and DART. The reported peak GPU memory is the average across 8 Nvidia A10 GPU}
    \label{tb:train}
\end{table}
\section{Experiments}
To validate the design of DART, we conduct extensive experiments and present the key results. Additional implementation details are provided in Appendices~\ref{sec:appendix:data}-\ref{sec:appendix:implementation}.
\begin{table}[!t]
    \centering
    \begin{tabular}{l c}
        \toprule
        Methods                   & Accuracy (\%) \\
        \midrule
        No-CoT                    & 32.5 \\
        DART                      & 42.6 \\
        ~~~~w/o ST                & 36.8 \\
        ~~~~w/o REM               & 36.2 \\
        ~~~~w/~~ LoRA-REM           & 40.8 \\
        ~~~~w/o $\loss_{distill}$ & 33.7 \\
        ~~~~w/ $\loss_{distill}$ on $y_1$    & 31.5 \\
        ~~~~w/ $\loss_{distill}$ on $[z_N;Y]$  & 33.7 \\
        \bottomrule
    \end{tabular}
    \caption{Ablation studies. \emph{LoRA-REM} indicates that we apply LoRA~\citep{hu2022lora} as REM.
    }
    \label{tb:ablation}
\end{table}

\subsection{Experimental Settings}

\textbf{Datasets.} Following previous work~\cite{deng2024icot,hao2024coconut}, we mainly fine-tune models on the complicated mathematical reasoning benchmark GSM8K-Aug~\citep{deng2023implicit}, an augmented version of GSM8K~\citep{Cobbe2021gsm}, which includes diverse math reasoning traces. To mitigate the risk of models memorizing the final answer from CoT sequences, the last step is omitted during training~\cite{shen2025codi}. 
For out-of-distribution evaluation, we adopt GSM-HARD~\citep{gao2023pal}, SVAMP~\citep{Patel2021are}, and MultiArith~\citep{Roy2015solving} as robustness benchmarks.

\textbf{Baselines.} We compare DART against three autoregressive methods and three non-autoregressive methods, namely: (1) CoT, which fine-tunes the model on CoT data to perform the traditional CoT reasoning; (2) Coconut~\citep{hao2024coconut}, which trains the model with CoT data in a mutil-stage manner and leverages autoregressively generated continuous thought; (3) CODI, similar to Coconut but employing a one-stage distillation framework using both CoT and no-CoT data; (4) No-CoT, which trains the model on no-CoT data to answer directly; (5)iCoT~\citep{deng2024icot}, which trains the model to reason in the hidden space by applying stepwise internaliztion; (6) and PauseFT~\citep{Goyal2024pause}, which inserts $C$ special filler tokens \texttt{<pause>} between the question and answer to allow extra computations. All methods are fine-tuned on Llama-3.2-1B~\citep{dubey2024llama} for consistency.

\begin{figure}[!t]
  \includegraphics[width=\columnwidth, trim=0 0 0 10pt, clip]{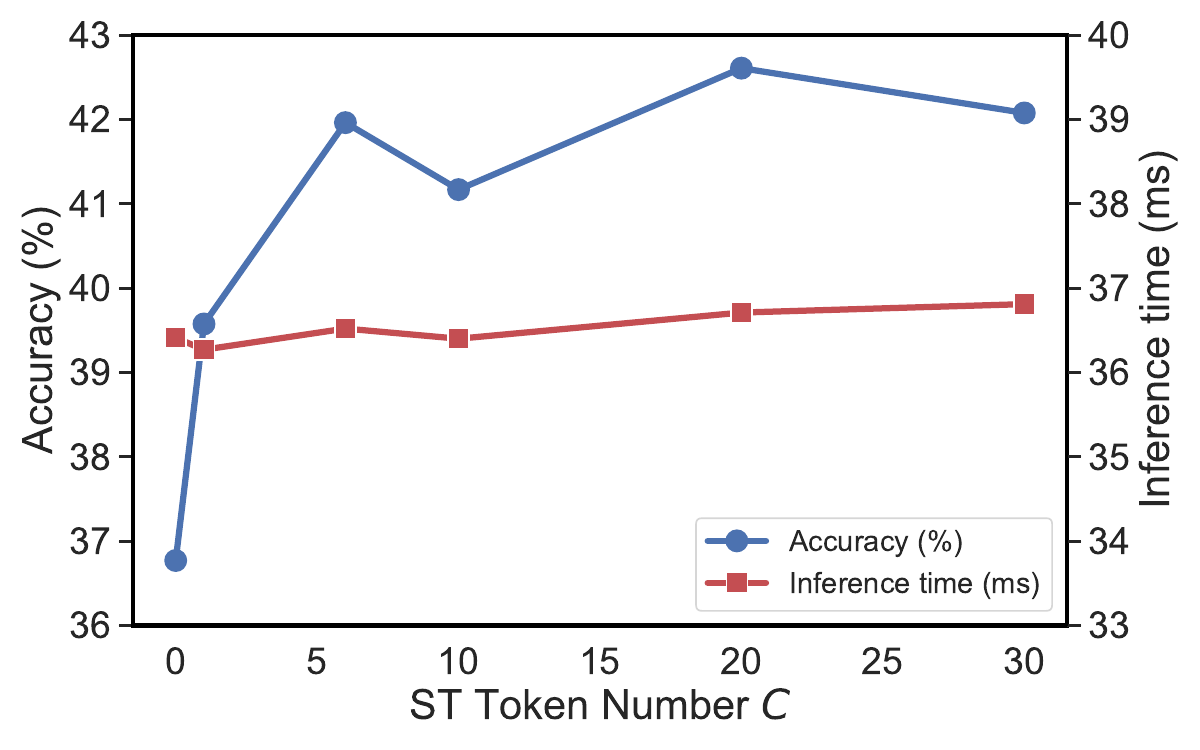}
  \caption{Accuracy and inference time on GSM8K with varying $C$, the number of ST tokens.}
  \label{fig:st_num}
\end{figure}

\subsection{Main Results}

\textbf{Comparison between Baselines.} Table~\ref{tb:main_res} summarizes the performance across GSM8K, GSM-HARD, SVAMP, and MultiArith.
To ensure a fair comparison of efficiency, we measure the inference time of all baselines on a Nvidia A10 GPU, even for those whose accuracy is sourced from prior reports. As shown, DART achieves the best performance among all NAR baselines on GSM8K, delivering a notable 10.1\% accuracy gain with negligible latency overhead (only 1 ms). On all out-of-distribution datasets, DART consistently outperforms other NAR methods, indicating robust generalization beyond the training distribution. While AR methods obtain higher accuracy, they suffer from significantly reduced inference efficiency due to stepwise generation, despite efforts to compress reasoning steps. These results demonstrate that DART achieves a compelling trade-off between accuracy and efficiency by fully leveraging distilled CoT knowledge in a single-step latent space. 
For training-compute transparency, we provide the total FLOPs and peak GPU memory for CoT, No-CoT, and DART in Table~\ref {tb:train}. All experiments were conducted on 8 Nvidia A10 GPUs. We can observe that DARt shares a comparable total FLOPs with the CoT baseline. Moreover, it's important to note the efficiency of DART's trainable parameters. For Llama-3.2-1B, CODI uses 98,574,336 trainable parameters, and Coconut employs a more complex multi-stage fully fine-tuning. In contrast, DART only has 44,040,192 trainable parameters, demonstrating an excellent efficiency trade-off considering both training and inference costs.

\textbf{Ablation Study.} To validate the contributions of key DART components, we evaluate the variants with certain components omitted or replaced. Our findings in Table~\ref{tb:ablation} are as follows: (1) Omitting $\loss_{distill}$, which transfers reasoning patterns from CoT trajectories, leads to a substantial performance drop; (2) Excluding the ST tokens impairs accuracy, likely due to the loss of CoT-derived positional priors; (3) Our proposed REM significantly enhances reasoning capability compared to using either no additional module or the vanilla LoRA; (4) Employing other tokens like the answer tokens as the distilling token in $\loss_{distill}$ hinders the effectiveness of alignment, empirically demonstrating the rationality of applying $\loss_{distill}$ on $z_N$.

\textbf{Sensitivity Analysis on ST Token Number.} We conduct the experiments on GSM8K with various values of the ST token number $C$. As shown in Figure~\ref{fig:st_num}, the accuracy rapidly increases during the initial stage as $C$ grows, then stabilizes once the token number becomes sufficient, demonstrating both the necessity and robustness of using ST tokens. Furthermore, benefiting from the non-autoregressive paradigm, DART introduces negligible latency even as $C$ increases.

\subsection{Further Analysis}
\label{sec:further_analysis}

\begin{table}[!t]
    \centering
    \resizebox{1.0\linewidth}{!}{
    \begin{tabular}{l c cc c cc}
        \toprule
         && \multicolumn{2}{c}{ProsQA}  && \multicolumn{2}{c}{CommonsenseQA}\\
        \cmidrule{3-4} \cmidrule{6-7}
        Methods && Acc & IT && Acc & IT \\
        \midrule
        \noalign{\smallskip}
        CoT
        && 98.8 & 882
        && 68.5 & 1471 \\
        No-CoT
        && 94.0 & 129
        && 65.4 & 34 \\
        DART 
        && 99.4 & 130
        && 75.1 & 34 \\
        \bottomrule
    \end{tabular}
    }
    \caption{Results on ProsQA and  CommonsenseQA-CoT. Acc and IT indicate the Accuracy (\%) and inference time (ms, on a Nvidia A10 GPU), respectively.
    }
    \label{tb:non_math}
\end{table}

\textbf{Task Generalizability.} To evaluate generalizability in other reasoning tasks, we fine-tune the models on the logic reasoning benchmark ProsQA~\citep{hao2024coconut} and commonsense reasoning benchmark CommonsenseQA-CoT~\cite{shen2025codi}. As summarized in Table~\ref{tb:non_math}, DART can even outperform the CoT baseline in ProsQA and CommonsenseQA-CoT while maintaining its low inference latency, demonstrating excellent robustness across diverse reasoning domains.

\textbf{Robustness across Various Models.} To further demonstrate the robustness of DART across different LLMs, we conduct the experiments using Llama-3.2-3B, Qwen2.5-1.5B~\citep{qwen2.5} and GPT2~\citep{radford2019language} as base models. Experimental results in Table~\ref{tb:models} show that DART can consistently boost the reasoning capabilities without notable latency across these models, demonstrating its scalability to larger and smaller models.

\begin{table}[!t]
    \centering
    \resizebox{0.93\linewidth}{!}{
    \begin{tabular}{l c cc c cc}
        \toprule
         && \multicolumn{2}{c}{No-CoT}  && \multicolumn{2}{c}{DART}\\
        \cmidrule{3-4} \cmidrule{6-7}
        Models && Acc & IT && Acc & IT \\
        \midrule
        \noalign{\smallskip}
        GPT2
        && 16.2 & 18 
        && 24.7 & 19 \\
        Qwen2.5-1.5B 
        && 33.3 & 95 
        && 42.2 & 95 \\
        Llama-3.2-3B 
        && 40.8 & 62 
        && 46.6 & 62 \\
        \bottomrule
    \end{tabular}
    }
    
    \caption{Results on GPT2, Qwen2.5-1.5B and Llama-3.2-3B. Acc and IT indicate the Accuracy (\%) and inference time (ms, on a Nvidia A10 GPU), respectively.
    }
    \label{tb:models}
\end{table}

\textbf{Qualitative Analysis.} We decode the final hidden states of ST tokens into natural language using the model’s word embeddings, finding that 69.9\% of the translated tokens match the words in ground-truth CoT. This ratio further rises to 80.1\% for cases that yield the correct answer. Additionally, we observe that the ST tokens also reflect a progressive reasoning process similar to CoT. For example, the top-1 ST translation is "\texttt{192 192 192 192 24 24 24 24 24 24 24 24 24 24 24 24 24 24 24 24}", corresponding to the CoT "\texttt{<<3*64=192>> <<192/8=24>>}".
\section{Conclusion}

We present DART, a fine-tuning framework that empowers LLMs to perform implicit reasoning in a non-autoregressive manner. By distilling knowledge from CoT data, the models trained with DART achieve a remarkable balance between accuracy and latency using the evolving ST tokens. Furthermore, extensive experiments confirm DART's robustness across various benchmarks and validate the effectiveness of its design.

\section*{Limitations}

\textbf{Additional Training Resources.} Due to its dual-pathway architecture, DART requires more computational resources. In our experiments, the proportion of trainable parameters is 2.86\%, compared to 1.79\% for LoRA.

\textbf{Supervision Signal.} Currently, DART aligns only the activation value of the last word before the answer between the CoT and ST pathways. This limited supervision may be suboptimal, as it can overlook some information in intermediate tokens. Incorporating more comprehensive supervision signals may help DART achieve better performance.

\textbf{Demand for CoT data.} DART relies on CoT data for knowledge distillation, which may not always be available. One potential solution is to use large LLMs capable of CoT reasoning to generate synthetic CoT data, though this approach incurs additional preprocessing costs.


\section*{Acknowledgments}

This work was supported by NSFC (62476123).


\bibliography{custom}

\begin{thebibliography}{26}
\providecommand{\natexlab}[1]{#1}

\bibitem[{Cheng and Durme(2024)}]{cheng2024compressed}
Jeffrey Cheng and Benjamin~Van Durme. 2024.
\newblock \href {https://doi.org/10.48550/ARXIV.2412.13171} {Compressed chain of thought: Efficient reasoning through dense representations}.
\newblock \emph{CoRR}, abs/2412.13171.

\bibitem[{Cobbe et~al.(2021)Cobbe, Kosaraju, Bavarian, Chen, Jun, Kaiser, Plappert, Tworek, Hilton, Nakano, Hesse, and Schulman}]{Cobbe2021gsm}
Karl Cobbe, Vineet Kosaraju, Mohammad Bavarian, Mark Chen, Heewoo Jun, Lukasz Kaiser, Matthias Plappert, Jerry Tworek, Jacob Hilton, Reiichiro Nakano, Christopher Hesse, and John Schulman. 2021.
\newblock \href {https://arxiv.org/abs/2110.14168} {Training verifiers to solve math word problems}.
\newblock \emph{CoRR}, abs/2110.14168.

\bibitem[{Dai et~al.(2023)Dai, Sun, Dong, Hao, Ma, Sui, and Wei}]{dai2023why}
Damai Dai, Yutao Sun, Li~Dong, Yaru Hao, Shuming Ma, Zhifang Sui, and Furu Wei. 2023.
\newblock \href {https://doi.org/10.18653/v1/2023.findings-acl.247} {Why can {GPT} learn in-context? language models secretly perform gradient descent as meta-optimizers}.
\newblock In \emph{Findings of the Association for Computational Linguistics}, pages 4005--4019.

\bibitem[{DeepSeek{-}AI et~al.(2025)DeepSeek{-}AI, Guo, Yang, Zhang, Song, Zhang, Xu, Zhu, Ma, Wang, Bi, Zhang, Yu, Wu, Wu, Gou, Shao, Li, Gao, Liu, Xue, Wang, Wu, Feng, Lu, Zhao, Deng, Zhang, Ruan, Dai, Chen, Ji, Li, Lin, Dai, Luo, Hao, Chen, Li, Zhang, Bao, Xu, Wang, Ding, Xin, Gao, Qu, Li, Guo, Li, Wang, Chen, Yuan, Qiu, Li, Cai, Ni, Liang, Chen, Dong, Hu, Gao, Guan, Huang, Yu, Wang, Zhang, Zhao, Wang, Zhang, Xu, Xia, Zhang, Zhang, Tang, Li, Wang, Li, Tian, Huang, Zhang, Wang, Chen, Du, Ge, Zhang, Pan, Wang, Chen, Jin, Chen, Lu, Zhou, Chen, Ye, Wang, Yu, Zhou, Pan, and Li}]{deepseek2025r1}
DeepSeek{-}AI, Daya Guo, Dejian Yang, Haowei Zhang, Junxiao Song, Ruoyu Zhang, Runxin Xu, Qihao Zhu, Shirong Ma, Peiyi Wang, Xiao Bi, Xiaokang Zhang, Xingkai Yu, Yu~Wu, Z.~F. Wu, Zhibin Gou, Zhihong Shao, Zhuoshu Li, Ziyi Gao, and 81 others. 2025.
\newblock \href {https://doi.org/10.48550/ARXIV.2501.12948} {Deepseek-r1: Incentivizing reasoning capability in llms via reinforcement learning}.
\newblock \emph{CoRR}, abs/2501.12948.

\bibitem[{Deng et~al.(2024)Deng, Choi, and Shieber}]{deng2024icot}
Yuntian Deng, Yejin Choi, and Stuart~M. Shieber. 2024.
\newblock \href {https://doi.org/10.48550/ARXIV.2405.14838} {From explicit cot to implicit cot: Learning to internalize cot step by step}.
\newblock \emph{CoRR}, abs/2405.14838.

\bibitem[{Deng et~al.(2023)Deng, Prasad, Fernandez, Smolensky, Chaudhary, and Shieber}]{deng2023implicit}
Yuntian Deng, Kiran Prasad, Roland Fernandez, Paul Smolensky, Vishrav Chaudhary, and Stuart~M. Shieber. 2023.
\newblock \href {https://doi.org/10.48550/ARXIV.2311.01460} {Implicit chain of thought reasoning via knowledge distillation}.
\newblock \emph{CoRR}, abs/2311.01460.

\bibitem[{Dubey et~al.(2024)Dubey, Jauhri, Pandey, Kadian, Al{-}Dahle, Letman, Mathur, Schelten, Yang, Fan, Goyal, Hartshorn, Yang, Mitra, Sravankumar, Korenev, Hinsvark, Rao, Zhang, Rodriguez, Gregerson, Spataru, Rozi{\`{e}}re, Biron, Tang, Chern, Caucheteux, Nayak, Bi, Marra, McConnell, Keller, Touret, Wu, Wong, Ferrer, Nikolaidis, Allonsius, Song, Pintz, Livshits, Esiobu, Choudhary, Mahajan, Garcia{-}Olano, Perino, Hupkes, Lakomkin, AlBadawy, Lobanova, Dinan, Smith, Radenovic, Zhang, Synnaeve, Lee, Anderson, Nail, Mialon, Pang, Cucurell, Nguyen, Korevaar, Xu, Touvron, Zarov, Ibarra, Kloumann, Misra, Evtimov, Copet, Lee, Geffert, Vranes, Park, Mahadeokar, Shah, van~der Linde, Billock, Hong, Lee, Fu, Chi, Huang, Liu, Wang, Yu, Bitton, Spisak, Park, Rocca, Johnstun, Saxe, Jia, Alwala, Upasani, Plawiak, Li, Heafield, Stone, and et~al.}]{dubey2024llama}
Abhimanyu Dubey, Abhinav Jauhri, Abhinav Pandey, Abhishek Kadian, Ahmad Al{-}Dahle, Aiesha Letman, Akhil Mathur, Alan Schelten, Amy Yang, Angela Fan, Anirudh Goyal, Anthony Hartshorn, Aobo Yang, Archi Mitra, Archie Sravankumar, Artem Korenev, Arthur Hinsvark, Arun Rao, Aston Zhang, and 82 others. 2024.
\newblock \href {https://doi.org/10.48550/ARXIV.2407.21783} {The llama 3 herd of models}.
\newblock \emph{CoRR}, abs/2407.21783.

\bibitem[{Feng et~al.(2023)Feng, Zhang, Gu, Ye, He, and Wang}]{feng2023towards}
Guhao Feng, Bohang Zhang, Yuntian Gu, Haotian Ye, Di~He, and Liwei Wang. 2023.
\newblock \href {http://papers.nips.cc/paper\_files/paper/2023/hash/dfc310e81992d2e4cedc09ac47eff13e-Abstract-Conference.html} {Towards revealing the mystery behind chain of thought: {A} theoretical perspective}.
\newblock In \emph{Advances in Neural Information Processing Systems 36}.

\bibitem[{Gao et~al.(2023)Gao, Madaan, Zhou, Alon, Liu, Yang, Callan, and Neubig}]{gao2023pal}
Luyu Gao, Aman Madaan, Shuyan Zhou, Uri Alon, Pengfei Liu, Yiming Yang, Jamie Callan, and Graham Neubig. 2023.
\newblock \href {https://proceedings.mlr.press/v202/gao23f.html} {{PAL:} program-aided language models}.
\newblock In \emph{International Conference on Machine Learning}, volume 202, pages 10764--10799.

\bibitem[{Goyal et~al.(2024)Goyal, Ji, Rawat, Menon, Kumar, and Nagarajan}]{Goyal2024pause}
Sachin Goyal, Ziwei Ji, Ankit~Singh Rawat, Aditya~Krishna Menon, Sanjiv Kumar, and Vaishnavh Nagarajan. 2024.
\newblock \href {https://openreview.net/forum?id=ph04CRkPdC} {Think before you speak: Training language models with pause tokens}.
\newblock In \emph{The Twelfth International Conference on Learning Representations}.

\bibitem[{Hao et~al.(2024)Hao, Sukhbaatar, Su, Li, Hu, Weston, and Tian}]{hao2024coconut}
Shibo Hao, Sainbayar Sukhbaatar, DiJia Su, Xian Li, Zhiting Hu, Jason Weston, and Yuandong Tian. 2024.
\newblock \href {https://doi.org/10.48550/ARXIV.2412.06769} {Training large language models to reason in a continuous latent space}.
\newblock \emph{CoRR}, abs/2412.06769.

\bibitem[{Hu et~al.(2022)Hu, Shen, Wallis, Allen{-}Zhu, Li, Wang, Wang, and Chen}]{hu2022lora}
Edward~J. Hu, Yelong Shen, Phillip Wallis, Zeyuan Allen{-}Zhu, Yuanzhi Li, Shean Wang, Lu~Wang, and Weizhu Chen. 2022.
\newblock \href {https://openreview.net/forum?id=nZeVKeeFYf9} {Lora: Low-rank adaptation of large language models}.
\newblock In \emph{The Tenth International Conference on Learning Representations}.

\bibitem[{Liu et~al.(2024)Liu, Liu, Zhou, and Ma}]{liu2024chain}
Zhiyuan Liu, Hong Liu, Denny Zhou, and Tengyu Ma. 2024.
\newblock \href {https://openreview.net/forum?id=3EWTEy9MTM} {Chain of thought empowers transformers to solve inherently serial problems}.
\newblock In \emph{The Twelfth International Conference on Learning Representations}.

\bibitem[{Loshchilov and Hutter(2019)}]{loshchilov2019adamw}
Ilya Loshchilov and Frank Hutter. 2019.
\newblock \href {https://openreview.net/forum?id=Bkg6RiCqY7} {Decoupled weight decay regularization}.
\newblock In \emph{7th International Conference on Learning Representations, {ICLR} 2019, New Orleans, LA, USA, May 6-9, 2019}. OpenReview.net.

\bibitem[{OpenAI(2025)}]{openai202501}
OpenAI. 2025.
\newblock \href {https://openai.com/index/learning-to-reason-with-llms/} {Learning to reason with llms}.

\bibitem[{Patel et~al.(2021)Patel, Bhattamishra, and Goyal}]{Patel2021are}
Arkil Patel, Satwik Bhattamishra, and Navin Goyal. 2021.
\newblock \href {https://doi.org/10.18653/v1/2021.naacl-main.168} {Are {NLP} models really able to solve simple math word problems?}
\newblock In \emph{Proceedings of the 2021 Conference of the North American Chapter of the Association for Computational Linguistics: Human Language Technologies}, pages 2080--2094.

\bibitem[{Radford et~al.(2019)Radford, Wu, Child, Luan, Amodei, and Sutskever}]{radford2019language}
Alec Radford, Jeff Wu, Rewon Child, David Luan, Dario Amodei, and Ilya Sutskever. 2019.
\newblock Language models are unsupervised multitask learners.

\bibitem[{Roy and Roth(2015)}]{Roy2015solving}
Subhro Roy and Dan Roth. 2015.
\newblock \href {http://arxiv.org/abs/1608.01413} {Solving general arithmetic word problems}.
\newblock In \emph{Proceedings of the 2015 Conference on Empirical Methods in NaturalLanguage Processing}, pages 1743--1752.

\bibitem[{Shao et~al.(2024)Shao, Wang, Zhu, Xu, Song, Zhang, Li, Wu, and Guo}]{shao2024deepseekmath}
Zhihong Shao, Peiyi Wang, Qihao Zhu, Runxin Xu, Junxiao Song, Mingchuan Zhang, Y.~K. Li, Y.~Wu, and Daya Guo. 2024.
\newblock \href {https://doi.org/10.48550/ARXIV.2402.03300} {Deepseekmath: Pushing the limits of mathematical reasoning in open language models}.
\newblock \emph{CoRR}, abs/2402.03300.

\bibitem[{Shen et~al.(2025)Shen, Yan, Zhang, Hu, Du, and He}]{shen2025codi}
Zhenyi Shen, Hanqi Yan, Linhai Zhang, Zhanghao Hu, Yali Du, and Yulan He. 2025.
\newblock \href {https://doi.org/10.48550/ARXIV.2502.21074} {{CODI:} compressing chain-of-thought into continuous space via self-distillation}.
\newblock \emph{CoRR}, abs/2502.21074.

\bibitem[{Sui et~al.(2025)Sui, Chuang, Wang, Zhang, Zhang, Yuan, Liu, Wen, Zhong, Chen, and Hu}]{sui2025stop}
Yang Sui, Yu{-}Neng Chuang, Guanchu Wang, Jiamu Zhang, Tianyi Zhang, Jiayi Yuan, Hongyi Liu, Andrew Wen, Shaochen Zhong, Hanjie Chen, and Xia~Ben Hu. 2025.
\newblock \href {https://doi.org/10.48550/ARXIV.2503.16419} {Stop overthinking: {A} survey on efficient reasoning for large language models}.
\newblock \emph{CoRR}, abs/2503.16419.

\bibitem[{Team(2024)}]{qwen2.5}
Qwen Team. 2024.
\newblock \href {https://qwenlm.github.io/blog/qwen2.5/} {Qwen2.5: A party of foundation models}.

\bibitem[{Wang et~al.(2024)Wang, Li, Shao, Xu, Dai, Li, Chen, Wu, and Sui}]{wang2024math}
Peiyi Wang, Lei Li, Zhihong Shao, Runxin Xu, Damai Dai, Yifei Li, Deli Chen, Yu~Wu, and Zhifang Sui. 2024.
\newblock \href {https://doi.org/10.18653/v1/2024.acl-long.510} {Math-shepherd: Verify and reinforce llms step-by-step without human annotations}.
\newblock In \emph{Proceedings of the 62nd Annual Meeting of the Association for Computational Linguistics (Volume 1: Long Papers)}, pages 9426--9439.

\bibitem[{Wei et~al.(2022)Wei, Wang, Schuurmans, Bosma, Ichter, Xia, Chi, Le, and Zhou}]{wei2022cot}
Jason Wei, Xuezhi Wang, Dale Schuurmans, Maarten Bosma, Brian Ichter, Fei Xia, Ed~H. Chi, Quoc~V. Le, and Denny Zhou. 2022.
\newblock \href {http://papers.nips.cc/paper\_files/paper/2022/hash/9d5609613524ecf4f15af0f7b31abca4-Abstract-Conference.html} {Chain-of-thought prompting elicits reasoning in large language models}.
\newblock In \emph{Advances in Neural Information Processing Systems 352}.

\bibitem[{Yu et~al.(2024)Yu, Jiang, Shi, Yu, Liu, Zhang, Kwok, Li, Weller, and Liu}]{yu2024metamath}
Longhui Yu, Weisen Jiang, Han Shi, Jincheng Yu, Zhengying Liu, Yu~Zhang, James~T. Kwok, Zhenguo Li, Adrian Weller, and Weiyang Liu. 2024.
\newblock \href {https://openreview.net/forum?id=N8N0hgNDRt} {Metamath: Bootstrap your own mathematical questions for large language models}.
\newblock In \emph{The Twelfth International Conference on Learning Representations}.

\bibitem[{Yue et~al.(2024)Yue, Qu, Zhang, Fu, Huang, Sun, Su, and Chen}]{yue2024mammoth}
Xiang Yue, Xingwei Qu, Ge~Zhang, Yao Fu, Wenhao Huang, Huan Sun, Yu~Su, and Wenhu Chen. 2024.
\newblock \href {https://openreview.net/forum?id=yLClGs770I} {Mammoth: Building math generalist models through hybrid instruction tuning}.
\newblock In \emph{The Twelfth International Conference on Learning Representations}.

\end{thebibliography}

\appendix

\section{Analysis of the Shift Value}
\label{sec:appendix:derivation}
\begin{table*}[!htbp]
    \centering
    \begin{tabular}{lp{0.73\linewidth}}
        \toprule
        Notation &  \makecell[c]{Mathematical Meaning}\\
        \midrule
        $l$ & The index of the current decoder layer, usually used as a superscript. \\
        $n, d$ & The hidden state dimension and the REM projection dimension. \\
        $\alpha$ & The hyperparameter for scaling in REM. \\
        $Q$ & The question sequence.\\
        $Y=\bbr{y_i}_{i=1}^{M}$ & The answer sequence of length $M$.\\
        $Z=\bbr{z_i}_{i=1}^{N}$ & The intermediate sequence of length $N$.\\
        $S=\bbr{s_i}_{i=1}^{C}$ & The ST sequence of length $C$, with each $s_i$ set as a special token \texttt{<st>}.\\
        $t$ & The index separating the CoT sequence $z_{1:t-1}$ and the separators $z_{t:N}$ in $Z$. \\
        $\mbr{\cdot;\cdot}$ & The concatenation operation for matrix pairs and sequence pairs.\\
        $X=\mbr{S;z_{t:N}}$ & The concatenation of the ST sequence and separators.\\
        $W_K^l,W_V^l$ & The key and value projection matrices. \\
        $W_{R1}^J,W_{R2}^J$ & The REM matrices for key projection matrices when $J=K$ and for value projection matrices when $J=V$. \\
        $H_Q^l,H_Z^l,H_X^l$ & The output hidden state matrices associated with $Q$, $Z$, and $X$. \\
        $\q^l$ & The attention query vector of the last separator $z_N$. \\
        $\a^l,\tilde{\a}^l,\widehat{\a}^l$ & The output vectors of the attention head corresponding to the last separator $z_N$ in the no-CoT, CoT, and ST cases. \\
        $\h^l,\tilde{\h}^l,\widehat{\h}^l$ & The output hidden states of the last separator $z_N$ corresponding to the no-CoT, CoT, and ST cases. \\
        $\theta,\phi$ & The parameters of LLM and REM.\\
        $\theta^{1:l},\phi^{1:l}$ & The parameters of the first $l$ layers of LLM and REM.\\
        $f(\cdot)$ & The feed-forward function in the decoder layer.\\
        $h_{\theta}(\cdot)$ & The generation function for the model to produce an answer sequence conditioned on the input. \\
        $g_{\theta^{1:l}}(\cdot),g_{\theta^{1:l},\phi^{1:l}}(\cdot)$ & The function for the model to produce a shift value conditioned on the input and the parameters in the subscript. \\
        \bottomrule
    \end{tabular}
    \caption{Frequently used notations along with their mathematical meaning.}
    \label{tb:notation}
\end{table*}

For better readability, we first summarize the mathematical notations adopted for this paper in Table~\ref{tb:notation}.

Similar to~\citep{dai2023why}, we mainly focus on the effect of the intermediate sequence 
$X$ on the hidden state of the last separator token $z_N$. Firstly, we can derive the simplified expression for the activation $\tilde{\a}^l$ of $z_N$ as follows:

\begin{align*}
    \tilde{\a}^l
    & = W_V^l[H_Q;H_{Z}]\text{softmax}\sbr{\frac{W_K^l[H_Q;H_{Z}]}{\sqrt{n}}}^T\q^l \\
    & \approx W_V^l[H_Q;H_{Z}]\sbr{W_K^l[H_Q;H_{Z}]}^T\q^l \\
    & = W_V^lH_Q\sbr{W_K^lH_Q}^T + W_V^lH_{Z}(W_K^lH_{Z})^T\q^l \\
    & \triangleq \a^l + W_V^lH_{Z}(W_K^lH_{Z})^T\q^l,
\end{align*}
where $W_K^l,W_V^l\in\R^{n\times n}$ are the key and value projection matrices of the $l$-th decoder layer, $H_{Q}$, $H_{Z}$ are the input hidden state of question $Q$ and intermediate tokens $Z$, $\q^l$ is the attention query vector corresponding to $z_{N}$, and $\a^l$ is the activation when only $Q$ is given. The superscript $l-1$ for $H_{Q}$, $H_{Z}$ is omitted for simplicity. The approximation in the second step is obtained by omitting the softmax operation and scaling factor $\sqrt{n}$. Then, by going through the feed-forward layer $f(\cdot)$, we can get the hidden state of $z_N$ in the $l$-th layer as follows:
\begin{equation*}
    \tilde{\h}^l \approx {\h}^l +f\sbr{W_V^lH_{Z}(W_K^lH_{Z})^T\q^l}
\end{equation*}
where $\h^l= f({\a}^l)$. Since $Z$ can be viewed as the output of LLM given $Q$, we further define
\begin{equation*}
    g_{\theta^{1:l}}\sbr{h_{\theta}\sbr{Q}} \triangleq f\sbr{W_V^lH_{Z}(W_K^lH_{Z})^T\q^l}.
\end{equation*}
Hence, the CoT effectively injects a shift in the hidden state of $z_N$, which can be parameterized by the model parameters $\theta$ and input $Q$. Based on this, we employ REM to construct $g_{\theta^{1:l},\phi^{1:l}}([Q;X])$ and apply an L1 distance loss to approximate this shift. 

\begin{table}[!htbp]
    \centering
    \begin{tabular}{l c c}
        \toprule
        Dataset     & Training  & Evaluation \\
        \midrule
        GSM8K-Aug   & 385620    & 1319 \\
        GSM-HARD    & -         & 1319 \\
        SVAMP       & -         & 1000 \\   
        MultiArith  & -         & 600 \\
        ProsQA      & 17886    & 500 \\
        CommonsenseQA-CoT & 8096& 1221 \\
        \bottomrule
    \end{tabular}
    \caption{Dataset statistics. GSM-HARD, SVAMP and MultiArith are only used for evaluation.}
    \label{tb:datasets}
\end{table}
\section{Datasets}
\label{sec:appendix:data}

\textbf{Statistics.} The statistics of utilized datasets are provided in Table~\ref{tb:datasets} 

\textbf{Examples} We provide some examples of the data used in our experiments.
\example{GSM8K-Aug}{"Andy receives a monthly salary of \$800 but he has to pay a tax of 7\%. How much is his net salary?"}{"<<800*7/100=56>> <<800-56=744>>"}{"744"}

\example{GSM-HARD}{"A robe takes 2287720 bolts of blue fiber and half that much white fiber. How many bolts in total does it take?"}{}{"3431580.0"}

\example{SVAMP}{"Each pack of dvds costs 76 dollars. If there is a discount of 25 dollars on each pack. How much do you have to pay to buy each pack?"}{}{"51.0"}

\example{MultiArith}{"Faye had 34 coloring books. If she gave away 3 of them, but then bought 48 more, how many would she have total?"}{}{"79"}

\example{ProsQA}{"Every yimpus is a yumpus. Jack is a yerpus. Max is a vumpus. Max is a zhorpus. Every brimpus is a rorpus. Every brimpus is a timpus. Every rorpus is a hilpus. Every gwompus is a yumpus. Every gorpus is a lempus. Every impus is a kerpus. Every impus is a brimpus. Every impus is a hilpus. Every kerpus is a boompus. Max is a gorpus. Every gorpus is a rempus. Every gorpus is a yimpus. Every gorpus is a rompus. Every yerpus is a timpus. Every kerpus is a rorpus. Every yerpus is a impus. Every gwompus is a yimpus. Every gorpus is a yumpus. Max is a gwompus. Every brimpus is a kerpus. Every gwompus is a vumpus. Every yumpus is a gerpus. Is Jack a zhorpus or boompus?"}{"Jack is a yerpus. Every yerpus is a impus. Every impus is a kerpus. Every kerpus is a boompus."}{"Jack is a boompus."}

\example{CommonsenseQA-CoT}{"The fox walked from the city into the forest, what was it looking for?
Choices: \\
A: pretty flowers. \\
B: hen house \\
C: natural habitat \\
D: storybook \\
E: dense forest"}{"The fox, being a wild animal, would typically seek its natural habitat when moving from the city into the forest. Options like "pretty flowers," "hen house," and "storybook" do not align with a fox's natural behavior, while "dense forest" describes the environment but not the fox's purpose. Therefore, "natural habitat" is the most logical choice."}{"C"}

\section{Implementation Details}
\label{sec:appendix:implementation}
For all experiments, we set $d=128$ and $\alpha=32$ for REM, consistent with the configuration used for LoRA~\citep{hu2022lora} to fine-tune the CoT pathway. We find that the distillation loss is much smaller than the other two losses. We set the trade-off hyperparameter $\lambda=20$ since we find that the distillation loss is much smaller than the other two losses. We employ the AdamW~\citep{loshchilov2019adamw} with a cosine annealing learning rate schedule following a 3\% warm-up period. Additionally, we enable the \texttt{bf16} in the trainer and evaluate the models using bfloat16 precision. 

We fine-tune the Llama-3.2-1B, the primary base model in our experiments, for 10 epochs with the learning rate initialized as 8e-4. For Llama-3.2-3B and Qwen2.5-1.5B, we initialize the learning rate at 5e-4 and keep all other configurations as those used for Llama-3.2-1B. For GPT2, we set the initial learning rate to 2e-3 and train the model for 40 epochs.

For Coconut, we adopt their official implementation and use the same number of training epochs as in our experiments.

\end{document}